\documentclass[runningheads]{llncs}
\usepackage{cite}
\usepackage{hyperref}
\usepackage[table,xcdraw]{xcolor}
\usepackage{multirow}
\usepackage{tabularx}
\usepackage{float}

\usepackage[T1]{fontenc}
\usepackage{graphicx}
\begin{document}
\title{Leveraging GPT-4o Efficiency for Detecting Rework Anomaly in Business Processes}
%
%
\author{Mohammad Derakhshan\inst{1}\orcidID{0009-0008-9028-0574} \and
Paolo Ceravolo\inst{1}\orcidID{0000-0002-4519-0173} \and
Fatemeh Mohammadi\inst{1}\orcidID{0000-0002-3100-7569}}
\authorrunning{M. Derakhshan \and et al.}
%
\institute{University of Milan, 20133 Milan, Italy}
\maketitle              
\begin{abstract}
This paper investigates the effectiveness of GPT-4o-2024-08-06, one of the Large Language Models (LLM) from OpenAI, in detecting business process anomalies, with a focus on \textit{rework anomalies}. In our study, we developed a GPT-4o-based tool capable of transforming event logs into a structured format and identifying reworked activities within business event logs.
The analysis was performed on a synthetic dataset designed to contain rework anomalies but free of loops. 
To evaluate the anomaly detection capabilities of GPT-4o-2024-08-06, we used three prompting techniques: \textit{zero-shot}, \textit{one-shot} and \textit{few-shot}. These techniques were tested on different anomaly distributions, namely \textit{normal}, \textit{uniform}, and \textit{exponential}, to identify the most effective approach for each case.
The results demonstrate the strong performance of GPT-4o-2024-08-06. On our dataset, the model achieved 96.14\% accuracy with \textit{one-shot} prompting for the \textit{normal} distribution, 97.94\% accuracy with \textit{few-shot} prompting for the \textit{uniform} distribution, and 74.21\% accuracy with \textit{few-shot} prompting for the \textit{exponential} distribution.
These results highlight the model's potential as a reliable tool for detecting rework anomalies in event logs and how anomaly distribution and prompting strategy influence the model's performance.
 
\end{abstract}
\section{Introduction}
Detecting anomalies in business processes is critical for ensuring operational efficiency, identifying fraud and improving overall process quality~\cite{Bohmer2016,Nolle2018,Barbon2020,Bellandi2021,Bena2024}. Traditional approaches have made significant progress, but face inherent limitations that reduce their effectiveness in real-world scenarios. For example, while Graph Neural Networks (GNNs) have been used to exploit object-centric event logs for anomaly detection \cite{b11}, they struggle to detect temporal order anomalies and often rely on complex graph structures that may not fully capture the conceptual connections between activities \cite{b2}. 
Conformance checking techniques excels at detecting control flow anomalies in Business Process Management (BPM), such as parallel execution or sequence rule violations, but struggles to incorporate multidimensional aspects such as resource usage or performance metrics~\cite{Dunzer2019}. Their extension to multidimensional aspects require significant manual configuration and domain expertise, which limits their flexibility in different scenarios~\cite{Carmona2022}.  
On the other hand, machine learning enables the detection of complex, multidimensional anomalies by analysing attributes such as task durations and resource workloads~\cite{Barbon2020}. However, it has difficulty preserving temporal order and enforcing process constraints, requiring additional transformations to handle relationships such as sequence or parallelism~\cite{Peeperkorn2023}.


Recently, Large Language Models (LLMs) have emerged as a powerful layer that simplifies the interface between users and BPM systems~\cite{Barbon2024, Ceravolo2024}. By leveraging natural language understanding and generation capabilities, LLMs enable non-expert users to interact more intuitively with BPM systems, reducing the complexity of traditional configurations and facilitating tasks such as anomaly detection, process mining and automation.
While existing LLM-based approaches such as DABL can effectively detect semantic anomalies, they often disrupt long-range dependencies and still require significant manual effort in trace generation and model training \cite{b5}.

By incorporating the conceptual and logical connections between activities and using the power of LLMs, specifically \textit{GPT-4o-2024-08-06}\cite{gpt-4o}, in this study, we developed an approach that enables a more intuitive and comprehensive detection of anomalies without the need for deep technical knowledge of the underlying business processes. This ease of use allows even general managers with limited technical expertise to apply the model efficiently. Additionally, unlike previous methods that rely on complex setups or costly implementation, this solution is simple, cost-effective, and offers actionable insights that enhance business decision-making. The ability to interpret anomalies in natural language further empowers businesses to understand not only when an anomaly occurs but also why, making this method a valuable tool for operational improvement.

This study aims to evaluate the performance of \textit{GPT-4o-2024-08-06} - hereafter referred to as \textit{GPT-4o} for simplicity - in rework anomaly detection using various prompt engineering techniques \cite{b2}, including \textit{zero-shot}, \textit{one-shot} and \textit{few-shot} prompts. By systematically adapting prompts and methods, this research seeks to identify the most effective strategies for using \textit{GPT-4o} in this context.
Three primary research questions guide the study:
(i) \textit{Is \textit{GPT-4o} effective in detecting rework anomaly in business processes?}
(ii) \textit{How does the distribution of anomalies affect the performance of GPT-4o?}
(iii) \textit{Which prompting technique performs best for different anomaly distributions?}
To address these questions, the experiments were structured around the distribution of anomalies within the event log. Specifically, we analysed three well-known statistical distributions: \textit{normal} \cite{normal_distribution}, \textit{uniform} \cite{uniform_distribution}, and \textit{exponential} \cite{exponential_distribution}.

The study demonstrates that utilizing the few-shot technique with a uniform distribution yields a 46\% higher accuracy compared to Principal Component Analysis, 30\% higher than Isolation Forest, and 16\% higher than the Deep Autoencoding Gaussian Mixture Model on the same dataset.

The paper is structured as follows. Section \ref{s:rw} reviews the related works, providing a comprehensive overview of prior research relevant to our study. Section \ref{s:meth} describes the methodology of our work. Section \ref{s:dataset} outlines the dataset used, including its characteristics and anomaly distribution. Section \ref{s:experiment} presents the experiment setup, explaining the configuration and execution of our study. Section \ref{s:results} illustrates the results and their significance. Section \ref{s:descution} elaborates on outcomes by highlighting the strengths and limitations and comparing the results to other machine learning models. Section \ref{s:future_works} highlights potential improvements and future works, offering directions for further research and refinement of the proposed methodology. Finally, section \ref{s:conclusion} summarizes our findings.

\section{Related works}\label{s:rw}
\paragraph{Conformance checking for anomaly detection in BPM.}
Anomaly detection in BPM has been effectively addressed using \textit{conformance checking} techniques~\cite{Dunzer2019}. These methods excel at detecting deviations from constraints explicitly defined by business process models, such as ensuring that certain activities are performed in parallel, or that prescribed preconditions are followed in a sequence preceding an executed task. For example, in a model where activities $A$ and $B$ must run concurrently, conformance checking can easily flag an sequence of events as non-compliant if $A$ and $B$ are not executed, but will not identify any anomaly based on the order of $A$ and $B$, which may follow any ordering. However, these techniques are inherently tied to the control-flow rules derived from the process model, making it difficult to incorporate insights from other event log dimensions, such as resource allocation, event's attributes or performance indicators. This limitation makes their applicability more challenging when dealing with anomalies that span multiple dimensions or features, often requiring domain-specific solutions~\cite{Carmona2022}.

\paragraph{Machine learning for multidimensional anomaly detection.}
Machine learning techniques have also been applied to anomaly detection in BPM, offering greater flexibility in capturing multidimensional aspects of event logs~\cite{Barbon2020}. For example, by analysing attributes such as resource workloads, task durations and case metadata, machine learning models can identify complex patterns of variation, such as cases where certain resources consistently take longer for certain tasks, which could indicate inefficiencies or skill mismatches~\cite{Pasquadibisceglie2021}. However, these techniques struggle with preserving the order of events and enforcing model-specific constraints, such as parallel activities or strict sequences. For example, recognising that $A$ must always occur before $B$ becomes less direct for a machine learning model, as these relationships can be encoded using specific methods for transforming the original data. As a result, while machine learning is powerful for multidimensional anomaly detection, it may not be able to enforce strict adherence to process structure, which is critical in many BPM applications~\cite{Peeperkorn2023}.

\paragraph{Object centric anomaly detection.}
One of the proposed approaches to business process anomaly detection is via graph neural networks on object-centric event logs~\cite{b2}. Unlike traditional process mining methods that rely on ``flattened'' event logs tied to a single case identifier, this approach exploits object-centric process mining where events can be associated with multiple cases. In~\cite{b2}  a graph convolutional autoencoder (GCNAE) architecture that reconstructs object-centric event logs as attributed graphs is exploited. Anomaly detection is performed by computing anomaly scores based on the reconstruction errors of these graphs, allowing the identification of anomalies at both activity and attribute levels.
Despite its innovations, the approach has notable limitations. A significant drawback is its poor performance in detecting anomalies related to the temporal sequence of events. Temporal order anomalies are critical in detecting anomalies in business processes, as many anomalies involve incorrect ordering of activities. Reliance on GNNs, which primarily aggregate local neighbourhood information, can obscure subtle temporal shifts, undermining the model's ability to effectively detect such patterns.
While the object-centric approach improves accuracy, it also introduces significant complexity \cite{object_centric_anomaly_detection_challenge}. Transforming object-centric process instances into graphs can be computationally expensive, especially for large datasets, potentially limiting scalability and efficiency in real-world applications with large event logs.
In addition, the model lacks a clear interpretability mechanism to explain the root causes of detected anomalies. This reduces its practical utility, as organisations often require actionable insights to effectively address and resolve problems in their processes.

\paragraph{Fuzzy based anomaly detection.}
In \cite{b3} a method for detecting anomalies in large event logs generated by Enterprise Resource Planning (ERP) systems is presented. The approach integrates three core techniques: process mining, fuzzy multi-attribute decision making, and fuzzy association rule learning.
Process mining is used to check the conformance of event logs to standard operating procedures and to identify deviations that may indicate anomalies.
Fuzzy multi-attribute decisioning assigns anomaly rates to these deviations based on expert judgement, enabling a nuanced assessment of their significance.
Fuzzy association rule learning generates rules to detect fraudulent behaviour by analysing anomalies at different confidence levels.
While this system provides a structured methodology, it has several limitations. One critical challenge is its heavy reliance on expert judgement to determine anomaly weights and rates. This reliance creates a bottleneck for real-world implementation, requiring ongoing expert input to configure and fine-tune fuzzy parameters and thresholds. The difficulty of optimising these thresholds across different business environments can lead to an overabundance of false positives or undetected anomalies, undermining the reliability of the system.

\paragraph{LLM based anomaly detection.}
One of the LLM-based anomaly detection approaches has been introduced by Wei Guan in~\cite{b5}. In this paper, the author presents a method that detects semantic anomalies in business processes. This approach leverages LLMs, specifically \textit{Llama 2}, fine-tuned using generated logs across various domains. 
Although this method improves semantic anomaly detection, it inherits a limitation common to other semantic-based approaches, namely the disruption of long-distance dependencies between events. Traces are split into event pairs, which can miss critical relationships throughout the process.
In addition, this approach depends heavily on simulated anomalies from the generated dataset. This might limit its effectiveness when applied to entirely novel datasets with unanticipated types of anomalies. The reliance on synthetic data could also result in gaps when addressing real-world, complex processes.

\paragraph{The proposed appraoch.}
In contrast to the approaches discussed, our study introduces a \textbf{GPT-4o}-based tool that leverages natural language processing to capture multidimensional aspects, minimize preprocessing efforts, reduce reliance on domain experts, and mitigate the complexity of traditional methods, making it an appealing solution for modern, dynamic BPM environments.

\section{Methodology}\label{s:meth}
This paper focuses on a specific type of anomaly in business processes known as rework. A \textit{rework anomaly} occurs when a task - or set of tasks - is unnecessarily repeated within the same process instance, often signaling inefficiencies that negatively impact overall process performance~\cite{b6}. Detecting rework anomalies is quite challenging because not all reworked activities follow the same pattern. For example, in the dataset analyzed in this paper, repeated activities may not occur consecutively but may be interspersed with other tasks, making it important to distinguish whether these repetitions serve a valid purpose or represent inefficiencies that should be classified as rework.

The study exploits the pattern recognition capabilities of LLMs~\cite{Mirchandani2023}, such as GPT-4o, to detect rework anomalies. Unlike traditional approaches, LLMs excel at interpreting nuanced patterns of activity, enabling them to distinguish between rework and intentional repetition by considering the broader context of process behavior.

One of the most significant weaknesses of LLMs is the limited size of the input. Rework anomalies were selected because they are self-contained, allowing the model to classify events without relying on variant-dependent analysis. This focus enables the model to detect inefficiencies independently, without the added complexity of analyzing process variant relationships and performing the analysis on a smaller chunk of data, unlike other types of anomalies, such as delay anomalies, that require considering the distribution of all activities to detect whether a specific activity is abnormal.

\section{Dataset}\label{s:dataset}
\begin{figure}
    \centering
    \includegraphics[width=0.3\textwidth]{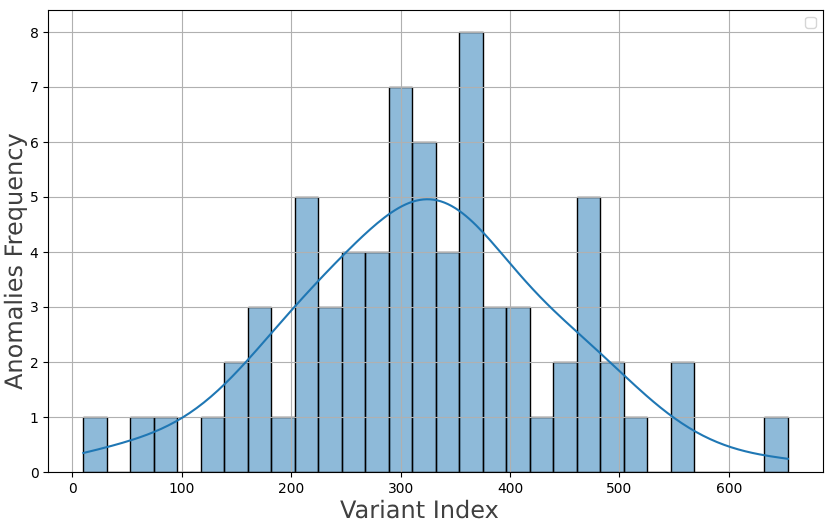}
    \hfill
    \includegraphics[width=0.3\textwidth]{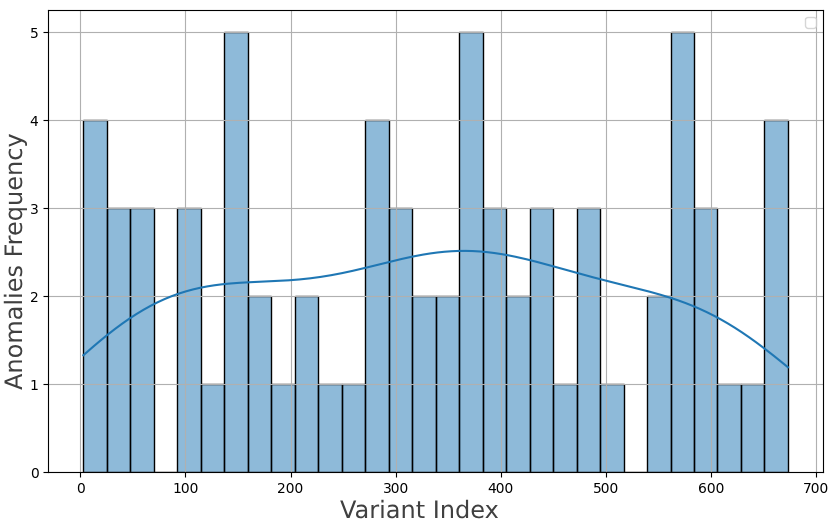}
    \hfill
    \includegraphics[width=0.3\textwidth]{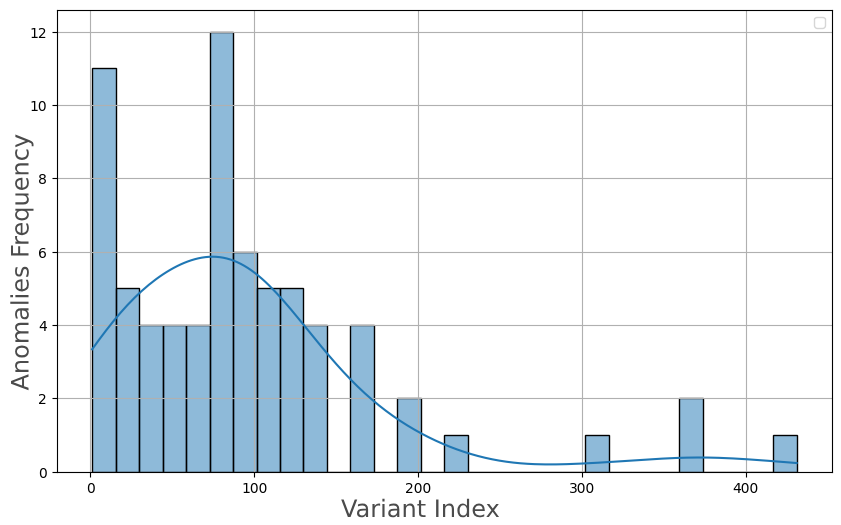}
    \caption{Anomaly distribution of datasets. From left to right: normal, uniform, and exponential distribution. The \textbf{x-axis} represents the \textbf{variant index} in the dataset, and the \textbf{y-axis} illustrates the \textbf{anomaly frequency}.}
    \label{fig:three_images}
\end{figure}

The paper uses a subset of the synthetic dataset presented in~\cite{dataset}. The original dataset contains 1,000 labelled variants, with labels indicating whether a variant is normal or has a rework anomaly. However, due to \textit{OpenAI}'s token-per-minute (TPM) limit for tier-1 users \cite{open_ai_max_token_per_min_limit}, we are limited to 30,000 tokens per minute. Consequently, we can only use a subset of this dataset, consisting of 760 variants, 71 reworked variants and 689 normal variants. We have created three copies of this dataset, each with a different anomaly distribution. The first distribution is normal. The second dataset uses a uniform distribution, and the third is based on an exponential distribution of anomalies. These three different distributions of anomalies are shown in Figure \ref{fig:three_images}.

For the normal distribution, the dataset is filtered to extract the normal instances, while the abnormal variants are separately identified. We generate a set of insertion indices to integrate these abnormal instances into the normal dataset. These indices are determined using a normal distribution centered around the midpoint of the normal dataset's length and scaled according to its variance. We used a {\tt Numpy} random number generator for uniform and exponential distribution, {\tt random.choice} and {\tt random.exponential} respectively.

\section{Experiment}\label{s:experiment}
To address the anomaly detection task in our study, we developed an application that uses LangChain~\cite{LangChain} as the backbone for workflow definition and execution. LangChain provides a modular and versatile framework that simplifies the complexity of building scalable, stateful and context-aware LLM-based applications. 
Our workflow is parameterised by four key messaging components. (i) \textit{Human message} represent user-initiated input to the system. (ii) \textit{System message} collects predefined, hard-coded messages that guide the LLM during execution by providing contextual instructions or constraints. (iii) \textit{AI message} refers to the outputs generated by the LLM based on the prompts and instructions provided. (iv) \textit{Function messages} are another form of system messages; however, their initiation differs from system messages and are used to describe functions.
The workflow consists of two main sub-processes: \textit{Setup} and \textit{LLM execution}.

\begin{figure}[h]
\centering
\includegraphics[scale=0.35]{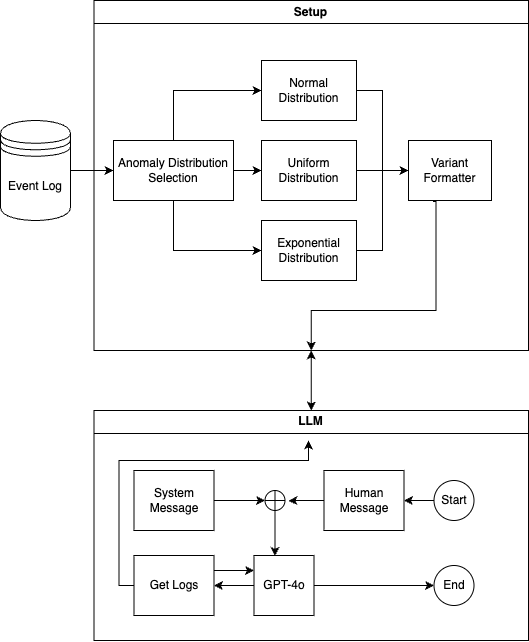}
\caption{Anomaly detection workflow}
\label{fig:figure2}
\end{figure}

\begin{itemize}
    \item \textit{Setup} sub-process: This stage involves preparing the input data for LLM analysis. Event logs are retrieved, a user-defined anomaly distribution is applied, and the logs are formatted into event variants compatible with the LLM. Table \ref{tab:table1} shows an example of such a formatted variant. 
    \item \textit{LLM execution} sub-process: In this phase, the system processes the event variants, integrates user prompts and provides specific instructions to the LLM to generate outputs. This sub-process ensures that the workflow runs smoothly and produces the desired results. The complete workflow is visualised in Figure \ref{fig:figure2}.
\end{itemize}

LangChain's architecture has allowed us to efficiently handle the complex interactions between these components, but also ensures that we can quickly adapt our solution in response to new challenges or technology updates. Moreover, it enables the creation of an application that is both easy to integrate with other systems and easy to use for the end user.

\begin{table}[ht]
\centering
\caption{Examples of formatted variants}\label{tab:table1}
\begin{tabular}{|l|l|}
\hline
Id         & Variant  \\ \hline
\#1  & Activity Q → Activity C → Activity R → Activity S \\ \hline
\#2  & Activity A → Activity A → Activity Z → Activity W  \\ \hline
\end{tabular}
\end{table}

The process begins by retrieving the \textit{Human message}, which is then combined with \textit{System messages} to initialise the LLM execution. At this stage the LLM assesses whether it needs to call the Setup sub-process. If so, it calls a function from the Setup subprocess to retrieve event variants. The LLM then processes these variants to perform the task requested by the user and returns the results.
The content of the \textit{human message} depends on the prompting method selected by the analyst, while the system messages remain consistent. Human messages are generated using a prompt template based on prompt engineering techniques. We used three prompt methods: (1) zero-shot, (2) one-shot, and (3) few-shot \cite{prompt_engineering}.
In the one-shot and few-shot examples, the rework pattern is guaranteed to be present in the dataset. The one-shot method contains a single example pattern, while the few-shot method contains three example patterns. To ensure the reliability of our results, we repeated each task three times. Table \ref{tab:table2} provides an overview of the messages used for each prompting technique.

\begin{table*}[]
\centering
\caption{Prompt utilization for each prompting method.}\label{tab:table2}
\resizebox{\textwidth}{!}{
\begin{tabular}{l|l|l|l|}
\cline{2-4}
                                                                          & \cellcolor[HTML]{EFEFEF}\textbf{Zero shot}                                                                                                                                                                                                                                    & \cellcolor[HTML]{EFEFEF}\textbf{One shot}                                                                                                                                                                                                                                                                                                                                                                                                                                                                                                 & \cellcolor[HTML]{EFEFEF}\textbf{Few shot}                                                                                                                                                                                                                                                                                                                                                                                                                                                                                                                                                                                                                                                                                                                                                                                                                                                                                                                                                                                                                                                                                                                                                                               \\ \hline
\multicolumn{1}{|l|}{\cellcolor[HTML]{EFEFEF}\textbf{System Message}}     & \begin{tabular}[c]{@{}l@{}}You are a business process analyst who is an anomaly\\ detection and event log analysis expert. Keep your\\ response short. Directly give the output response;\\ there is no need for additional information.\end{tabular}                         & \begin{tabular}[c]{@{}l@{}}You are a business process analyst who is an anomaly\\ detection and event log analysis expert. Keep your\\ response short. Directly give the output response;\\ there is no need for additional information.\end{tabular}                                                                          & \begin{tabular}[c]{@{}l@{}}You are a business process analyst who is an anomaly\\ detection and event log analysis expert. Keep your\\ response short. Directly give the output response;\\ there is no need for additional information.\end{tabular}                                                                                                  \\ \hline
\multicolumn{1}{|l|}{\cellcolor[HTML]{EFEFEF}\textbf{Human Message}}      & \begin{tabular}[c]{@{}l@{}}Detect the variants with the rework anomalies.\\ The output should be in this format:\\ Variant Id\# Which sequence of activities is reworked?\\ Example:1\# A-\textgreater{}A-\textgreater{}B-\textgreater{}B-\textgreater{}A\end{tabular}        & \begin{tabular}[c]{@{}l@{}}Detect the variants with the rework anomalies.\\ An example of a rework anomaly is:\\ Activity Q -\textgreater Activity C -\textgreater Activity R -\textgreater Activity S -\textgreater\\ Activity G -\textgreater Activity X -\textgreater Activity X\\ Because it has two repetitions of activity X in a row.\\ The output should be in this format:\\ Variant Id\# Which sequence of activities is reworked?\\ Example:1\# A-\textgreater{}A-\textgreater{}B-\textgreater{}B-\textgreater{}A\end{tabular} & \begin{tabular}[c]{@{}l@{}}Detect the variants with the rework anomalies.\\ Three examples of rework anomalies are:\\ 1) Activity Q -\textgreater Activity C -\textgreater Activity R -\textgreater Activity S\\ -\textgreater Activity G -\textgreater Activity X -\textgreater Activity X -\textgreater Activity I\\ Because it has two repetitions of activity X in a row.\\ 2) Activity B -\textgreater Activity Q -\textgreater Activity C -\textgreater Activity Q\\ -\textgreater Activity C -\textgreater Activity G -\textgreater Activity X -\textgreater Activity T\\ -\textgreater Activity O \\ Because it has two sequence repetitions of Activity Q\\ -\textgreater Activity C, exactly in a row\\ 3) Activity Q -\textgreater Activity C -\textgreater Activity R -\textgreater Activity S\\ -\textgreater Activity C -\textgreater Activity R -\textgreater Activity G -\textgreater Activity C\\ -\textgreater Activity R\\ It has three repetitions of Activity C -\textgreater Activity R\\ in a row. \\ The output should be in this format:\\ Variant Id\# Which sequence of activities is reworked?\\ Example:1\# A-\textgreater{}A-\textgreater{}B-\textgreater{}B-\textgreater{}A\end{tabular} \\ \hline
\multicolumn{1}{|l|}{\cellcolor[HTML]{EFEFEF}\textbf{Function Message }} & \begin{tabular}[c]{@{}l@{}}This function generates the process event variants.\\ Each variant has an id indicated by an integer number\\ followed by a \# symbol. Each variant contains an\\ ordered sequence of activities separated by a -\textgreater symbol.\end{tabular} & \begin{tabular}[c]{@{}l@{}}This function generates the process event variants.\\ Each variant has an id indicated by an integer number\\ followed by a \# symbol. Each variant contains an\\ ordered sequence of activities separated by a -\textgreater symbol.\end{tabular}                                                                               & \begin{tabular}[c]{@{}l@{}}This function generates the process event variants.\\ Each variant has an id indicated by an integer number\\ followed by a \# symbol. Each variant contains an\\ ordered sequence of activities separated by a -\textgreater symbol.\end{tabular}\\
\hline
\end{tabular}
}
\vspace{-5pt}
\noindent\begin{flushleft}\fontsize{7}{7}\selectfont
\end{flushleft}
\end{table*}

\begin{figure}[h]
\centering
\includegraphics[scale=0.50]{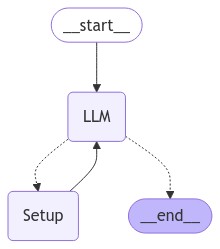}
\caption{Anomaly detection state graph}
\label{fig:figure3}
\end{figure}

The application is orchestrated using LangGraph \cite{b8}. It facilitates the integration of tool invocation capabilities into the LLM, which seamlessly aligns with our goals. Using LangGraph, GPT-4o can dynamically decide whether to invoke the Setup sub-process to handle user requests or rely on existing event logs and information to generate a response. The graph-based implementation of this approach is shown in Figure \ref{fig:figure3}.

We apply the above prompting techniques to each anomaly distribution dataset and repeat it three times. We then measure the performance of the system according to four criteria: \textit{Precision}, \textit{Recall}, \textit{Accuracy} and \textit{F1-Score}. The final reported result for each method in each scenario is an average of three runs. For each run, we waited 5 minutes for the model to finish. Beyond this time, we considered the inability of the model to analyse the anomalies and provide a response, resulting in zero performance, as we can see for \textit{zero-shot} prompting when we have an exponential distribution of anomalies. Our results are summarised in Table \ref{tab:table3}. The best performance for each criterion is highlighted in green.

\begin{table*}[]
\centering
\caption{Performance of GPT-4o for Anomaly detection for each prompting method on different anomaly distribution} \label{tab:table3}
\resizebox{\textwidth}{!}{%
\begin{tabular}{|c|c|c|c|c|c|c|c|c|c|c|c|c|c|c|}
\hline
\rowcolor[HTML]{D9D9D9} 
Distribution                  & Prompt type                 & Times & TP & TN  & FP  & FN & Precision (\%) & Recall (\%) & F1 score (\%) & Accuracy (\%) & Avg Precision (\%)                                    & Avg recall (\%)                                       & Avg F1 Score (\%)                                     & Avg Accuracy (\%)                               \\ \hline
                              &                             & 1     & 59 & 653 & 36  & 12 & 62,11          & 83,1        & 71,08         & 93,68         &                                                       &                                                       &                                                       &                                                 \\ \cline{3-11}
                              &                             & 2     & 63 & 667 & 22  & 8  & 74,12          & 88,73       & 80,77         & 96,05         &                                                       &                                                       &                                                       &                                                 \\ \cline{3-11}
                              & \multirow{-3}{*}{Zero Shot} & 3     & 56 & 661 & 28  & 15 & 66,67          & 78,87       & 72,26         & 94,34         & \multirow{-3}{*}{67,63}                         & \multirow{-3}{*}{83,56}                         & \multirow{-3}{*}{74,70}                         & \multirow{-3}{*}{94,69}                         \\ \cline{2-15} 
                              &                             & 1     & 66 & 657 & 32  & 5  & 67,35          & 92,96       & 78,11         & 95,13         &                                                       & \cellcolor[HTML]{DAF2D0}                              & \cellcolor[HTML]{DAF2D0}                              & \cellcolor[HTML]{DAF2D0}                        \\ \cline{3-11}
                              &                             & 2     & 61 & 662 & 27  & 10 & 69,32          & 85,92       & 76,73         & 95,13         &                                                       & \cellcolor[HTML]{DAF2D0}                              & \cellcolor[HTML]{DAF2D0}                              & \cellcolor[HTML]{DAF2D0}                        \\ \cline{3-11}
                              & \multirow{-3}{*}{One Shot}  & 3     & 60 & 686 & 3   & 11 & 95,24          & 84,51       & 89,55         & 98,16         & \multirow{-3}{*}{77,30}                         & \multirow{-3}{*}{\cellcolor[HTML]{DAF2D0}87,79} & \multirow{-3}{*}{\cellcolor[HTML]{DAF2D0}81,46} & \multirow{-3}{*}{\cellcolor[HTML]{DAF2D0}96,14} \\ \cline{2-15} 
                              &                             & 1     & 48 & 670 & 19  & 23 & 71,64          & 67,61       & 69,57         & 94,47         & \cellcolor[HTML]{DAF2D0}                              &                                                       &                                                       &                                                 \\ \cline{3-11}
                              &                             & 2     & 47 & 688 & 1   & 24 & 97,92          & 66,2        & 78,99         & 96,71         & \cellcolor[HTML]{DAF2D0}                              &                                                       &                                                       &                                                 \\ \cline{3-11}
\multirow{-9}{*}{Normal}      & \multirow{-3}{*}{Few Shot}  & 3     & 49 & 688 & 1   & 22 & 98             & 69,01       & 89,99         & 96,97         & \multirow{-3}{*}{\cellcolor[HTML]{DAF2D0}89,18} & \multirow{-3}{*}{67,60}                         & \multirow{-3}{*}{79,51}                         & \multirow{-3}{*}{96,05}                         \\ \hline
                              &                             & 1     & 48 & 684 & 5   & 23 & 90,57          & 67,61       & 77,44         & 96,32         &                                                       &                                                       &                                                       &                                                 \\ \cline{3-11}
                              &                             & 2     & 50 & 682 & 7   & 21 & 87,72          & 70,42       & 78,05         & 96,32         &                                                       &                                                       &                                                       &                                                 \\ \cline{3-11}
                              & \multirow{-3}{*}{Zero Shot} & 3     & 52 & 682 & 7   & 19 & 88,14          & 73,24       & 79,95         & 96,58         & \multirow{-3}{*}{88,81}                               & \multirow{-3}{*}{70,42}                         & \multirow{-3}{*}{78,48}                               & \multirow{-3}{*}{96,40}                   \\ \cline{2-15} 
                              &                             & 1     & 56 & 688 & 1   & 15 & 98,95          & 78,94       & 87,35         & 97,89         & \cellcolor[HTML]{DAF2D0}                              &                                                       &                                                       &                                                 \\ \cline{3-11}
                              &                             & 2     & 52 & 688 & 1   & 19 & 98,11          & 73,24       & 83,9          & 97,37         & \cellcolor[HTML]{DAF2D0}                              &                                                       &                                                       &                                                 \\ \cline{3-11}
                              & \multirow{-3}{*}{One Shot}  & 3     & 56 & 688 & 1   & 15 & 98,25          & 78,94       & 87,06         & 97,89         & \multirow{-3}{*}{\cellcolor[HTML]{DAF2D0}98,43} & \multirow{-3}{*}{77,04}                               & \multirow{-3}{*}{86,10}                         & \multirow{-3}{*}{97,71}                   \\ \cline{2-15} 
                              &                             & 1     & 59 & 687 & 2   & 12 & 96,72          & 83,099      & 89,58         & 98,03         &                                                       & \cellcolor[HTML]{DAF2D0}                              & \cellcolor[HTML]{DAF2D0}                              & \cellcolor[HTML]{DAF2D0}                        \\ \cline{3-11}
                              &                             & 2     & 65 & 685 & 4   & 6  & 94,2           & 91,55       & 92,87         & 98,68         &                                                       & \cellcolor[HTML]{DAF2D0}                              & \cellcolor[HTML]{DAF2D0}                              & \cellcolor[HTML]{DAF2D0}                        \\ \cline{3-11}
\multirow{-9}{*}{Uniform}     & \multirow{-3}{*}{Few Shot}  & 3     & 55 & 683 & 6   & 16 & 90,16          & 77,61       & 83,33         & 97,11         & \multirow{-3}{*}{93,69}                         & \multirow{-3}{*}{\cellcolor[HTML]{DAF2D0}84,08} & \multirow{-3}{*}{\cellcolor[HTML]{DAF2D0}88,59} & \multirow{-3}{*}{\cellcolor[HTML]{DAF2D0}97,94} \\ \hline
                              &                             & 1     & 0  & 0   & 0   & 0  & 0              & 0           & 0             & 0             &                                                       &                                                       &                                                       &                                                 \\ \cline{3-11}
                              &                             & 2     & 0  & 0   & 0   & 0  & 0              & 0           & 0             & 0             &                                                       &                                                       &                                                       &                                                 \\ \cline{3-11}
                              & \multirow{-3}{*}{Zero Shot} & 3     & 0  & 0   & 0   & 0  & 0              & 0           & 0             & 0             & \multirow{-3}{*}{0}                                   & \multirow{-3}{*}{0}                                   & \multirow{-3}{*}{0}                                   & \multirow{-3}{*}{0}                             \\ \cline{2-15} 
                              &                             & 1     & 60 & 455 & 234 & 11 & 20,41          & 84,51       & 32,83         & 67,76         &                                                       &                                                       &                                                       &                                                 \\ \cline{3-11}
                              &                             & 2     & 62 & 178 & 511 & 9  & 10,82          & 87,32       & 19,15         & 31,58         &                                                       &                                                       &                                                       &                                                 \\ \cline{3-11}
                              & \multirow{-3}{*}{One Shot}  & 3     & 54 & 445 & 244 & 17 & 18,12          & 76,06       & 29,17         & 65,66         & \multirow{-3}{*}{16,45}                               & \multirow{-3}{*}{82,63}                               & \multirow{-3}{*}{27,05}                               & \multirow{-3}{*}{55}                            \\ \cline{2-15} 
                              &                             & 1     & 61 & 596 & 93  & 10 & 39,61          & 85,92       & 54,22         & 86,45         & \cellcolor[HTML]{DAF2D0}                              & \cellcolor[HTML]{DAF2D0}                              & \cellcolor[HTML]{DAF2D0}                              & \cellcolor[HTML]{DAF2D0}                        \\ \cline{3-11}
                              &                             & 2     & 60 & 416 & 273 & 11 & 18,02          & 84,51       & 29,78         & 62,63         & \cellcolor[HTML]{DAF2D0}                              & \cellcolor[HTML]{DAF2D0}                              & \cellcolor[HTML]{DAF2D0}                              & \cellcolor[HTML]{DAF2D0}                        \\ \cline{3-11}
\multirow{-9}{*}{Exponential} & \multirow{-3}{*}{Few Shot}  & 3     & 61 & 498 & 191 & 10 & 24,21          & 85,92       & 37,68         & 73,55         & \multirow{-3}{*}{\cellcolor[HTML]{DAF2D0}27,28}       & \multirow{-3}{*}{\cellcolor[HTML]{DAF2D0}85,45}       & \multirow{-3}{*}{\cellcolor[HTML]{DAF2D0}40,56}       & \multirow{-3}{*}{\cellcolor[HTML]{DAF2D0}74,21} \\ \hline
\end{tabular}%
}
\end{table*}

\section{Results}\label{s:results}

We compared the performance of three prompting techniques across synthetic data set with different distributions, to assess their effectiveness in anomaly detection. Our findings suggest that GPT-4o demonstrates strong performance in anomaly detection tasks on event logs, although certain limitations and areas for improvement have emerged from our experiments. More specifically, this study was guided by addressing three primary research questions.

First, \textit{Are LLMs, such as GPT-4o, effective at detecting anomalies in business processes?} Our findings indicate that GPT-4o demonstrates strong performance in detecting anomalies, achieving a precision of 98.43\% when provided with a single example in a uniform distribution. This highlights GPT-4o’s capacity to detect anomalies accurately, even with minimal business context.

Second, \textit{Does the anomaly distribution affect the performance of LLMs in anomaly detection}? Our results confirm that the anomaly distribution significantly affects the performance of the model. In particular, the LLMs performed best under a uniform distribution, achieving high precision, accuracy and F1 scores. In contrast, performance declined under an exponential distribution, where anomalies were concentrated at the beginning of the event log (see Figure \ref{fig:three_images}). This skewed distribution reduced the overall effectiveness of the model, highlighting the challenges that LLMs such as GPT-4o may face when dealing with highly unbalanced anomaly distributions. 

Third, \textit{Which prompting technique performs best across different anomaly distributions?} We compared three prompting techniques - \textit{zero-shot}, \textit{one-shot} and \textit{few-shot} - across our datasets. Our analysis shows that in most cases the \textit{few-shot} technique produced the best results across anomaly distributions, particularly in terms of precision. However, this method also led to trade-offs: while the few-shot prompt achieved high precision, it recorded the lowest recall metric under a normal distribution, as shown in Table \ref{tab:table3}. Providing more examples to the model may reduce generalisation by forcing the model to focus on the examples\cite{zero-vs-few}. Therefore, the choice of prompting technique is critical in balancing \textit{precision} and \textit{recall}, depending on the anomaly distribution.

\section{Discussion}\label{s:descution}

GPT-4o demonstrated promising results overall. As shown in Table \ref{tab:table3}, its performance is highly influenced by the anomaly distribution. For instance, precision values range widely from 27.28\% to 98.43\%. The model achieves its best overall performance under a \textit{uniform distribution} and its worst under an \textit{exponential distribution}. These findings are consistent with previous observations regarding the unique characteristics of data distributions in event logs, which often deviate from the assumptions of standard machine learning models~\cite{Ceravolo2024}.

The choice of prompting strategy has a significant impact on whether precision or recall is prioritised. For example, for a \textit{uniform distribution}, the highest precision is achieved with \textit{one-shot} prompting, while the highest recall is achieved with \textit{few-shot} prompting. A similar pattern is observed for a \textit{normal distribution}, where the highest precision is obtained with few-shot prompting, while the best recall is obtained with one-shot prompting. This variability highlights the importance of tailoring prompting techniques to specific evaluation priorities.

For a comparative analysis, we incorporate results from Y. Shi et al. \cite{shi2024}, who evaluated the same dataset using traditional machine learning methods such as Local Outlier Factor (LOF), Principal Component Analysis (PCA), Isolation Forest (IF), Deep Autoencoding Gaussian Mixture Model (DAGMM), and TVPM, an original method proposed in~\cite{shi2024}. These methods provide a benchmark for anomaly detection accuracy and false discovery rate. However, a significant limitation in the comparative results lies in the authors' lack of consideration for anomaly distribution. The comparison between the best performance of each distribution and above methods, is shown in Table \ref{tab:table4}.

\begin{table}[]
\centering
\caption{Performance of GPT 4o compared to other Machine Learning methods}
\label{tab:table4}
\begin{tabular}{|l|c|l|}
\hline
Method               & \multicolumn{1}{l|}{Accuracy} & FDR  \\ \hline
LOF                  & 53\%                          & 46\% \\ \hline
PCA                  & 51\%                          & 55\% \\ \hline
IF                   & 67\%                          & 59\% \\ \hline
DAGMM                & 81\%                          & 75\% \\ \hline
TVPM                 & 87\%                          & 89\% \\ \hline
GPT 4o - Normal      & 96\%                          & 24\% \\ \hline
GPT 4o - Uniform     & 97\%                          & 9\%  \\ \hline
GPT 4o - Exponential & 74\%                          & 75\% \\ \hline
\end{tabular}
\end{table}

A key observation from Table \ref{tab:table4} is the notable difference in False Discovery Rate (FDR) between GPT-4o and other methods. Specifically, GPT-4o achieves an FDR of 9\% under a uniform distribution, significantly lower than the various approaches, highlighting how accurately GPT-4o was able to detect anomalies.

However, our study has also highlighted the limitations of GPT-4o for the detection of anomalies in business processes. A primary limitation is the API's token limit, which restricts the amount of data (in this case, business process variants) that can be entered in a single prompt. Tokens represent words, character sets or combinations of words and punctuation that LLMs use to parse text \cite{b9}. Like most LLMs, GPT-4o has a maximum token limit \cite{b16}, which restricts the amount of data that can be processed at one time. As a result, we have provided data in incremental requests, which can lead to inconsistencies or missed connections between pieces of data, potentially affecting overall performance.

%
%
%
%

\section{Future Works}\label{s:future_works}

While GPT-4o demonstrated strong performance in detecting rework anomalies, its applicability to other anomalies remains untested in our study. Rework anomalies are only a subset of the potential inefficiencies that can occur in business processes. Other types of anomalies, such as timing or sequence anomalies, may require different detection approaches, and it is unclear how well GPT-4o would perform in these cases without further experimentation.

Another important consideration is the refinement of prompt engineering. In our study, we experimented with \textit{zero-shot}, \textit{one-shot} and \textit{few-shot} prompting techniques, but future research could investigate more sophisticated prompt engineering strategies. For example, dynamic prompting - where the model is progressively exposed to more complex examples based on its performance - could improve its generalisation capabilities while maintaining high accuracy across different datasets.

Another area of improvement could be to improve the model's ability to handle negations. Negation tasks are particularly challenging for LLMs, but advances in model tuning and adversarial training could help address this issue. Exposing the model to more diverse training data, including nuanced negation examples, could improve its performance on such tasks, leading to more accurate anomaly detection, especially in cases where no anomaly is present.

Another promising area for future research is the integration of GPT-4o with other anomaly detection frameworks. For example, combining GPT-4o with traditional machine learning models such as random forests or neural networks could lead to hybrid systems that exploit the strengths of both approaches. These systems could balance interpretability, scalability and accuracy, making them more suitable for business environments.

Ultimately, integrating LLMs into business process anomaly detection is a promising step towards more intelligent and adaptive systems that can help organisations streamline operations, reduce inefficiencies and make better decisions based on data-driven insights.

\section{Conclusion}\label{s:conclusion}

In this study, we explored the potential of GPT-4o as an advanced business process anomaly detection tool, explicitly targeting rework anomalies. We demonstrated the robustness of the model in identifying inefficiencies, while highlighting the importance of tailored prompt engineering, using \textit{zero-shot}, \textit{one-shot} and \textit{few-shot} prompt techniques on datasets with different anomaly distributions. Our results demonstrate that GPT-4o, with its high accuracy and low false discovery rates under optimal conditions, is a valuable addition to process mining and anomaly detection methods.

However, the study also highlights GPT-4o's challenges in handling complex distributions, such as the exponential anomaly distribution, and scaling to larger datasets due to token limitations. These limitations point to the need for further advances in prompt engineering and hybrid systems that integrate LLMs with traditional machine learning techniques. Furthermore, the variability in performance between prompt strategies highlights the importance of tailoring method selection to specific dataset characteristics and detection priorities.

Overall, this research highlights the effectiveness of GPT-4o as an accessible tool for anomaly detection in dynamic business environments. By bridging the gap between technical and non-technical users, GPT-4o enables accurate detection and promotes deeper insight into the nature of anomalies. Future research into broader anomaly types, improved prompting strategies and hybrid frameworks promise to extend its applicability and further enhance its capabilities.

%
%
%
%

\end{document}